\documentclass{article}
\usepackage{graphicx} 
\usepackage[T1]{fontenc}
\usepackage{amsmath}
\usepackage{amssymb} 
\usepackage{authblk}
\usepackage{comment}
\usepackage{url}

\title{SCORE: A Semantic Evaluation Framework for Generative Document Parsing}

\author[1]{Renyu Li\thanks{email: \texttt{leah@unstructured.io}; Corresponding author}}
\author[1]{Antonio Jimeno Yepes}
\author[1]{Yao You}
\author[1]{Kamil Pluciński}
\author[1]{Maximilian Operlejn}
\author[1]{Crag Wolfe}

\affil[1]{Unstructured Technologies \protect\\ \texttt{https://unstructured.io}}

\date{}

\begin{document}
\maketitle

\begin{abstract}
Multi-modal generative document parsing systems challenge traditional evaluation: unlike deterministic OCR or layout models, they often produce semantically correct yet structurally divergent outputs. Conventional metrics---CER, WER, IoU, or TEDS---misclassify such diversity as error, penalizing valid interpretations and obscuring system behavior.

We introduce \textbf{SCORE} (\textit{Structural and COntent Robust Evaluation}), an interpretation-agnostic framework that integrates (i) adjusted edit distance for robust content fidelity, (ii) token-level diagnostics to distinguish hallucinations from omissions, (iii) table evaluation with spatial tolerance and semantic alignment, and (iv) hierarchy-aware consistency checks. Together, these dimensions enable evaluation that embraces representational diversity while enforcing semantic rigor.

Across \textbf{1,114 pages} spanning a holistic benchmark  and a field dataset, SCORE consistently revealed cross-dataset performance patterns missed by standard metrics. In \textbf{2--5\% of pages} with ambiguous table structures, traditional metrics penalized systems by \textbf{12--25\% on average}, leading to distorted rankings. SCORE corrected these cases, recovering equivalence between alternative but valid interpretations. Moreover, by normalizing generative outputs into a format-agnostic representation, SCORE reproduces traditional scores (e.g., table F1 up to \textbf{0.93}) without requiring object-detection pipelines, demonstrating that generative parsing alone suffices for comprehensive evaluation.

By exposing how interpretive diversity impacts evaluation outcomes and providing multi-dimensional, interpretable diagnostics, SCORE establishes foundational principles for semantically grounded, fair, and practical benchmarking of modern document parsing systems.
\end{abstract}

\section{Introduction}

Document parsing~\footnote{Document Processing: \url{https://en.wikipedia.org/wiki/Document_processing}} constitutes a fundamental component of intelligent document processing, enabling applications ranging from automated invoice extraction to retrieval-augmented generation (RAG)~\cite{lewis2020retrieval} systems. Traditional document parsing architectures employ deterministic pipelines that sequentially combine optical character recognition (OCR), layout analysis, and rule-based table extraction to produce structured outputs. The evaluation of these systems has relied on well-established task-specific metrics including Character Error Rate (CER) and Word Error Rate (WER)~\cite{kanai1993performance,neudecker2021survey}, Intersection-over-Union (IoU)~\cite{everingham2010pascal,lin2014microsoft}, and Tree Edit Distance-based Similarity (TEDS)~\cite{zhong2020image}. These metrics operate under the assumption of unique ground truth representations, rewarding exact matches while systematically penalizing any structural deviations.

The emergence of multi-modal generative document parsing systems has fundamentally transformed this landscape. Vision Language Models (VLMs) such as GPT-5 Mini, Gemini 2.5 Flash, and Claude Sonnet 3.7/4 \cite{gpt5mini2025, gemini25flash2025, claude37sonnet2025, claude4sonnet2025}, generate holistic document interpretations that integrate visual, textual, and structural signals in an end-to-end manner. Unlike their deterministic predecessors, these systems frequently produce outputs that are \emph{semantically correct yet structurally divergent}. Consider a table containing merged cells: one system may represent it as a flattened token sequence preserving reading order, while another generates hierarchical HTML markup with explicit structural relationships. Both interpretations faithfully capture the semantic content, yet traditional evaluation frameworks treat them as fundamentally incompatible, systematically misclassifying valid alternative interpretations as parsing errors.

This mismatch of the evaluation paradigm has significant practical implications. Current metrics conflate distinct aspects of parsing quality--semantic fidelity, hallucination control, and structural reasoning--into rigid string-level or tree-level comparisons. This conflation distorts system rankings, penalizes legitimate interpretive flexibility, and obscures consistent performance patterns across datasets. Without appropriate evaluation methodologies, researchers cannot make meaningful comparisons between generative and deterministic approaches, while practitioners face substantial uncertainty when selecting models for production deployment.

We address these fundamental limitations through \textbf{SCORE} (\textit{Structural and COntent Robust Evaluation}), an interpretation-agnostic evaluation framework specifically designed for multi-modal generative document parsing systems. \\
SCORE advances the state of evaluation in four critical dimensions: (1) \textit{adjusted edit distance} for a robust evaluation of content fidelity that tolerates structural reorganization, (2) \textit{token-level diagnostics} that separate content omissions from hallucinations, (3) \textit{table evaluation} incorporating semantic alignment and spatial tolerance for legitimate structural variations, and (4) \textit{hierarchy-aware consistency} assessment for document structure understanding. These innovations collectively embrace representational diversity while maintaining rigorous semantic accuracy standards.

Across 1,114 pages from two datasets, SCORE revealed reliable behavioral patterns that traditional metrics systematically obscured. Our evaluation encompasses 6 document parsing solutions offered by the Unstructured platform \cite{unstructured2025}: 4 VLM-based approaches (GPT-5 Mini, Gemini 2.5 Flash, and Claude Sonnet 3.7/4) and 2 deterministic pipelines ---the open-source and proprietary hi\_res strategy. On 2--5\% of pages with alternative table interpretations, raw NED scores diverged by 12--25\%, distorting rankings by making GPT-5 Mini appear weaker than Gemini 2.5 Flash (0.888 vs.\ 0.891) despite near equivalence once adjusted (0.896 vs.\ 0.895). TEDS exhibited similar inconsistencies, reversing system rankings across datasets due to sensitivity to representational variance rather than extraction quality. By contrast, SCORE's adjusted metrics yield better cross-dataset rankings, identifying Gemini~2.5 as the highest-fidelity system (Adj.\ NED 0.895/0.883) while revealing system-specific trade-offs: GPT-5 Mini records the lowest hallucination rates (TokensAdded 0.043/0.039), with traditional OCR-based methods trailing in second place as anticipated. Its content fidelity, however, remains inconsistent across datasets. Sonnet~4 emphasizes structural accuracy over coverage. This demonstrates that interpretation-agnostic evaluation avoids misleading penalties and exposes stable system behaviors that single-number metrics systematically miss.

\paragraph{Contributions.} This work makes the following contributions:
\begin{itemize}
    \item We identify and characterize four fundamental limitations of current document parsing evaluation (deterministic assumptions, format rigidity, semantic blindness, spatial inflexibility) and demonstrate their practical consequence: ranking distortion through empirical analysis across multiple systems and datasets.
    \item We introduce \textbf{SCORE}, a comprehensive interpretation-agnostic evaluation framework integrating adjusted edit distance, token-level diagnostics, semantic table evaluation, and hierarchy-aware consistency assessment.
    \item We demonstrate that SCORE corrects distorted rankings while enabling fine-grained diagnostics, reproducing classical detection metrics without requiring explicit object detection pipelines.
\end{itemize}
By reframing evaluation to recognize interpretive diversity while enforcing semantic rigor, SCORE establishes a principled foundation for fair, semantically grounded benchmarking of modern document parsing systems.
\section{Related Work}

This work focuses on the evaluation of document parsing systems, in which a document is analyzed to generate a structured output. This structured output might be used for several purposes, from extracting the total from an invoice to generating output suitable for a RAG system. The scope is further limited to the structure parsing correctness and extraction coherence considering the original document, this means as well that metrics for document summarization such as ROUGE or BLEU, or Visual Question Answering, are not specifically relevant to this work. On the other hand, we would like to identify ways in which the relevant metrics might provide relevant insights for error analysis.

Currently, there is a broad set of methods for performing document parsing. There are more traditional methods that rely on pipeline methods that concatenate different components with specific tasks, e.g. document layout understanding, OCR for text extraction and specific methods for structuring tables and/or figures, for instance. Also, there are end-to-end methods that rely on vision language models. Independently of the system being used in document parsing, we would like to be able to measure the capabilities of such systems in structuring documents.

In this section, we present a set of existing metrics that have been considered to evaluate document parsing capabilities. We categorize existing evaluation approaches into four main areas: layout analysis assessment, text extraction metrics (e.g. OCR evaluation), table extraction-specific evaluation, and evaluation of the structure of the document. Also, we enumerate existing benchmarks for document parsing and the annotation they contain for the existing metrics, including emerging vision-language model benchmarks.

\subsection{Document Layout Analysis Metrics}

Many traditional systems rely on a first stage in the processing pipeline in which the different layout elements are identified, which might include title, paragraphs, lists, among others and the coordinates of the enclosing bounding box. Document layout analysis evaluation has primarily adopted computer vision object detection metrics, particularly IoU and mean Average Precision (mAP). The PASCAL VOC and COCO detection challenges established evaluation protocols using mAP across multiple IoU thresholds, providing comprehensive assessment of localization accuracy \cite{everingham2010pascal,lin2014microsoft}. IoU measures the overlap between predicted and ground truth bounding boxes, with typical thresholds of 0.5 determining true positive classifications:

\begin{equation}
\text{IoU} = \frac{\text{Area of Intersection}}{\text{Area of Union}}
\end{equation}

These metrics excel at evaluating spatial precision but fail to assess content semantic correctness or interpretive variations in document structure understanding and have been used in research work and many community challenges in which these metrics have been used,e.g.~\cite{jimeno2021icdar}. However, these evaluation benchmarks enforce rigid sets of labels and penalize systems for reasonable interpretations of the alternative layout.

\subsection{Text Extraction Metrics}

In traditional systems, text is extracted from documents either from the document itself when possible or relies on the processing by an OCR system. Current VLMs are capable of processing the image of a document page using a neural network without having to apply segmentation approaches to find the characters. Independently of the method used to extract text from a document, there are interests in identifying how well the text has been extracted from a document. The scope of evaluation reflects the whole page of the layout elements identified in the document.

Traditional OCR evaluation has established standardized metrics centered on character- and word-level accuracy measurements. The most widely adopted metrics are CER and WER, both grounded in Levenshtein distance calculations \cite{kanai1993performance,neudecker2021survey}. CER measures the percentage of characters incorrectly recognized, calculated as:

\begin{equation}
\text{CER} = \frac{\text{Insertions} + \text{Deletions} + \text{Substitutions}}{\text{Total Characters in Ground Truth}}
\end{equation}

WER operates analogously at the word level. Several normalization approaches to these two metrics have been proposed~\cite{marzal1993edit}, as shown in equation~\ref{eq:ned}, where $s$ is the predicted text, $g$ is the ground truth reference text and $L_{\text{edit}}$ is the Levenshtein distance.

\begin{equation}
  \text{NED}(s, g) = 1 - \min\left(\max\left(\frac{L_{\text{edit}}(s, g)}{\max(|g|, |s|)}, 0\right), 1\right)
  \label{eq:ned}
\end{equation}
  
Established benchmarks suggest that good OCR accuracy corresponds to CER of 1-2\% for printed text, with rates greater than 10\% considered poor \cite{holley2009good}. the International Conference on Document Analysis and Recognition (ICDAR) competitions have been instrumental in standardizing these evaluation protocols, particularly through frameworks such as the ISRI OCR evaluation tools that became part of established OCR evaluation frameworks~\cite{santos2019ocr}.

However, these metrics suffer from fundamental limitations when applied to generative document parsing. They assume deterministic, character-exact outputs and fail to accommodate semantic equivalence across different but valid representations. A system that correctly interprets "Table 1" as "Tab. 1" receives an identical penalty to one that completely misrecognizes the content, despite vastly different semantic implications.

Furthermore, VLMs come with the risk of hallucination (see~\cite{huang2025hallucination} for a survey of hallucination metrics), which implies in addition to erroneous text extraction output, the addition of output not present in the original document. Existing OCR metrics measure different types of error in a single metric, making it challenging to properly perform an error analysis for these new models. This also implies more variability in the more deterministic output of more traditional document intelligence methods, which overwhelms traditional metrics.

\subsection{Table Structure Evaluation}

Tables are an important layout element, since a large amount of information in documents is structured in tabular format. Once properly identified in the document, tables might be structured for better processing. This implies identifying the cells of the table and the content of each cell. Specific representations have been used, including using mark-up and HTML that model the structure and content.

Table parsing evaluation has seen significant advancement through the introduction of TEDS \cite{zhong2020image}. TEDS addresses the multi-dimensional nature of table understanding by modeling tables as hierarchical tree structures and computing normalized edit distances between predicted and ground truth representations. TEDS is defined in equation~\ref{eq:teds}, where $T_{a}$ and $T_{b}$ are the tables being compared, $EditDist$ is the tree distance. TEDS captures structural relationships and cell alignments more effectively than simple text-based metrics. Additional metrics exist such as S-TEDS~\cite{huang2023CVPR}, which evaluates only the structure of the table.

\begin{equation}
TEDS(T_{a}, T_{b}) = 1 - \frac{EditDist(T_{a}, T_{b})}{max(|T_{a}|, |T_{b}|)}
\label{eq:teds}
\end{equation}

Recent work on large-scale table recognition datasets and challenged, such as TabRecSet with 38.1K tables \cite{yang2023large}, PubTabNet~\cite{zhong2020image} and ICDAR challenged~\cite{jimeno2021icdar}, has promoted TEDS as the standard for table evaluation. The metric has been adopted across major benchmarks and demonstrates superior alignment with human judgments compared to character-level metrics. However, TEDS still operates under the assumption of single correct interpretations, penalizing alternative but semantically equivalent table structures.

There are additional table evaluation metrics, which employ detection-based metrics borrowed from object detection, including precision, recall, and F-scores for table detection, combined with cell-level content accuracy assessments~\cite{gilani2017table,hassan2010towards,smock2022pubtables}. These approaches, while more nuanced than pure OCR metrics, remain constrained by rigid ground truth matching requirements.

\subsection{Document structure evaluation}

The final dimension in which document parsing systems can be evaluated would be to order the extracted elements in a way that follows what could be defined as reading order. In more traditional systems, this would imply applying algorithms such as XY-cut~\cite{ha1995recursive,meunier2005optimized} once the layout elements have been identified. Current VLMs are capable of providing this in a holistic way. For this task, specialized metrics based on Levenshtein distance and pairwise ordering accuracy have been developed to assess a system's ability to correctly sequence text blocks. Advanced methods further incorporate penalty matrices and information retrieval concepts such as precision and recall to provide a comprehensive framework for this crucial task~\cite{clausner2013significance,zhang2024modeling}. As we can imagine, several valid interpretations of the reading order of more complex documents might exist, and thus these metrics might be too strict in some cases.

\subsection{Document Parsing Benchmark Evolution}

Historically, document parsing benchmarks are task-specific and progress was driven by competitions and datasets that focused on isolated problems. For example, ICDAR has long featured contests on specific tasks like text localization and segmentation in well-defined domains, such as historical maps and scanned magazines.

Similarly, foundational data sets such as RVL-CDIP~\cite{harley2015evaluation} became the standard for document classification, but a critical evaluation revealed significant flaws, including an estimated 8.1\% label error rate, ambiguous annotations, and data leakage between training and test sets. These issues, combined with a persistent reliance on rigid, exact-match metrics like the F-score , created a systemic inadequacy where models could perform well on a benchmark while failing to demonstrate true, generalizable understanding. This has been partly solved by larger benchmark sets focused on several aspects of document parsing~\cite{zhong2020image, smock2022pubtables, jimeno2021icdar, pfitzmann2022doclaynet}.

Recent benchmark development has recognized the limitations of traditional evaluation settings and metrics, leading to the development of more comprehensive evaluation frameworks. OmniDocBench \cite{ouyang2025omnidocbench}, accepted to CVPR 2025, represents the most ambitious effort to date, featuring 1,000 real-world documents from nine sources, including academic papers, financial reports, newspapers, and handwritten notes. The benchmark provides localization information for 15 block-level and 4 span-level document elements, with over 20K and 80K annotations respectively.

OmniDocBench supports multi-level evaluation across 19 layout categories and 15 attribute labels, enabling flexible assessment from end-to-end performance to task-specific analysis. However, despite its comprehensive coverage, OmniDocBench continues to rely fundamentally on exact matching between predicted and ground truth annotations, inheriting the limitations of traditional evaluation approaches.

DocBench \cite{zou2024docbench} specifically targets VLMs, evaluating systems like GPT-4 and Claude-3 on document understanding tasks. The benchmark revealed interesting performance patterns: Claude-3 excels in metadata comprehension while GPT-4 shows limitations in unanswerable question categories. However, DocBench evaluation remains constrained by predetermined correct answers, limiting its ability to recognize alternative valid interpretations.

Complementary OCR-focused benchmarks like OmniAI ocr-benchmark \cite{getomniAI2025ocrbenchmark} address specific challenges including rotated layouts, charts, and diagrams, while CC-OCR \cite{yang2024cc} focuses on evaluating large multi-modal models in literacy tasks. These specialized benchmarks, while valuable, continue to employ traditional accuracy-based metrics that penalize interpretive diversity.

In addition to traditional ICDAR competitions, there are also recent developments in leaderboards~\cite{IDPLeaderboard} that periodically evaluate recent visual large language models using the metrics described above, which allows comparing the output of these models on existing evaluation corpora. The metrics and benchmarks considered in these leaderboards will suffer from the same issues mentioned in this section.

\subsection{Limitations of Current Approaches}

Despite significant advances in benchmark development and metric sophistication presented above, current evaluation approaches share fundamental limitations that our work addresses:

\textbf{Deterministic Assumption}: Existing metrics assume single correct outputs, failing to recognize that document parsing, particularly for ambiguous or complex layouts, naturally yields multiple valid interpretations.

\textbf{Format Rigidity}: Traditional approaches cannot meaningfully compare outputs across different representational formats (HTML vs. plain text vs. structured JSON), limiting comprehensive system comparison.

\textbf{Semantic Blindness}: Character and structure-based metrics miss semantic equivalence, penalizing systems that preserve meaning while varying in presentation.

\textbf{Spatial Inflexibility}: Current metrics lack tolerance for minor spatial variations that preserve semantic meaning, particularly problematic for table and layout evaluation.

These limitations create systematic biases in evaluation, potentially misguiding research directions and deployment decisions. Our interpretation-agnostic framework directly addresses these challenges by embracing the interpretive diversity inherent in modern generative document parsing systems.

\section{Methodology}

  We introduce an interpretation-agnostic evaluation framework designed to address a central challenge in multi-modal generative document parsing: how to fairly evaluate systems that may generate
  semantically correct but representationally divergent outputs. Unlike traditional deterministic evaluation, our framework explicitly embraces representational diversity while maintaining rigorous accuracy
   standards.

  \subsection{Core Evaluation Principles}

  Generative document parsing systems frequently exhibit legitimate interpretive variability: two systems may represent identical semantic content using different but equally valid strategies. Our framework
   addresses this through three principles:

  \begin{itemize}
      \item \textbf{Semantic Equivalence Recognition.} Representations preserving semantic meaning are treated as equivalent, emphasizing information preservation over surface-level fidelity.
      \item \textbf{Spatial Tolerance.} Minor positional variations that do not affect interpretation are ignored.
      \item \textbf{Structural Flexibility.} Logical relationships may be validly represented using different hierarchical structures.
  \end{itemize}

\subsection{Content Fidelity Assessment}

  We begin with content-level evaluation as the foundation of document parsing assessment, building upon established edit distance practices while introducing normalization and structural interpretation
  solutions.

  \subsubsection{Adjusted NED for Structural Alignment}

  Traditional concatenated edit distance fails when semantically equivalent content is organized differently due to legitimate structural interpretation. For example, consider a quarterly revenue table:

\begin{center}
\begin{tabular}{|l|l|}
\hline
Q1 & \$100K \\
Q2 & \$200K \\
\hline
\end{tabular}
\end{center}

One system may extract it as a flattened sequence:

\begin{quote}
\texttt{Q1 \$100K Q2 \$200K}
\end{quote}

while another outputs column-wise grouping:

\begin{quote}
\texttt{Q1 Q2 \$100K \$200K}
\end{quote}

Both preserve the same key–value pairs, yet concatenated edit distance treats them as divergent, producing 
artificially low similarity scores. We introduce \textbf{Adjusted NED} using word-weighted fuzzy alignment across table and non-table elements:

  \begin{equation}
  \text{NED}_{\text{adj}}(s, g) = \max \left( \text{NED}(s, g), \frac{\sum_{k \in \mathcal{K}} W_k}{W_{\text{total}}} \right),
  \end{equation}

  where $W_k = \sum_{i \in \text{elements}(k)} w_i \cdot \text{Sim}_k(e_i, g)$ with element-specific similarity functions:  fuzzy matching for tables, standard NED for paragraphs, and
  caption-level alignment for figures.

  \subsubsection{Reading Path Diagnostics}

  To differentiate between information loss and alternative reading paths,  we use complementary token-level frequency-aware metrics. The \textbf{\textit{TokensFound}} metric measures the proportion of reference tokens successfully preserved, independent of ordering, thereby identifying content loss. In contrast, the \textbf{\textit{TokensAdded}} metric measures the proportion of spurious tokens generated by the system. This serves as a direct indicator of \emph{hallucination level}, since higher values imply the introduction of unsupported or extraneous content. We operate on multisets (bags) of tokens that preserve frequency
  information:

  \begin{equation}
  \text{TokensFound}(s, g) = \frac{\sum_{t} \min(\text{freq}_s(t), \text{freq}_g(t))}{\sum_{t} \text{freq}_g(t)},
  \end{equation}

  \begin{equation}
  \text{TokensAdded}(s, g) = \frac{\sum_{t} \max(0, \text{freq}_s(t) - \text{freq}_g(t))}{\sum_{t} \text{freq}_s(t)}.
  \end{equation}

  where $\text{freq}_s(t)$ and $\text{freq}_g(t)$ represent the frequency of token $t$ in system output $s$ and ground truth $g$ respectively. For example, if ground truth tokens are \{Q1, \$100K, Q2, \$200K\} and system output is \{Q1, Q2, \$100K, \$300K, \$100K\}, then:
  \begin{itemize}
      \item TokensFound = $\frac{1+1+1+0}{1+1+1+1} = 0.75$ (missing \$200K)
      \item TokensAdded = $\frac{0+0+1+1}{1+1+2+1} = 0.40$ (extra \$100K and hallucinated \$300K)
  \end{itemize}
 
\subsection{Table Evaluation Framework}

Evaluating tables in generative outputs requires more than spatial overlap. Two systems may segment headers or cell boundaries differently yet still preserve identical semantics. To account for this, our framework moves beyond IoU-style metrics, emphasizing semantic equivalence, structural flexibility, and interpretable diagnostics. We begin by normalizing diverse output formats into a unified representation, enabling consistent, fair, and format-agnostic comparison.

\subsubsection{Format-Agnostic Representation}

Outputs in HTML, JSON, or structured text are first mapped into a common semantic form that captures content, structure, and relationships:

\begin{itemize}
    \item \textbf{HTML:} \texttt{<td>Q1</td><td>\$100K</td>} $\;\rightarrow$ direct cell extraction  
    \item \textbf{JSON:} \texttt{\{row:0, col:0, content:"Q1"\}} $\;\rightarrow$ metadata mapping  
    \item \textbf{Custom:} \texttt{\{"x":0, "y":0, "w":1, "h":1, "content":"Q1"\}} \\
    $\;\rightarrow$ coordinate-based parsing  
\end{itemize}

All formats normalize into equivalent tuples such as  
\texttt{(row=0, col=0, "Q1")}, \texttt{(row=0, col=1, "\$100K")}.  
This process, also applicable to Markdown and other structured formats, enables the \textbf{\textit{CellAlignment}} function to compare cells using both content similarity and positional correspondence with built-in spatial tolerance.

\subsubsection{Four Innovations Beyond Traditional Metrics}

Our framework introduces four innovations over spatial-overlap baselines:

\paragraph{Semantic-First Detection.}
We recast table detection as a content-based classification problem rather than a boundary-matching task, scoring with the F-measure:
\begin{equation}
F_\beta = \frac{(1 + \beta^2) \cdot P \cdot R}{\beta^2 \cdot P + R}.
\end{equation}
This ensures that tables are matched through content similarity, avoiding penalties for boundary interpretation differences.
It is especially critical for VLMs, which lack explicit object detection modules, rendering IoU metrics inapplicable. By focusing on semantic agreement rather than bounding-box overlap, this approach enables fair evaluation across both VLMs and deterministic systems.

\paragraph{Granular Cell-Level Analysis with Spatial Tolerance.}
We separate \emph{content accuracy} (text recognition) from \emph{index accuracy} (spatial relationships) and evaluate alignments under positional shifts $\delta \in \{-N,\ldots,N\}$:
\begin{equation}
\text{ContentAcc}(P, G), \text{IndexAcc}(P, G) = \max_{\delta \in \{-N,\ldots,N\}} \text{CellAlignment}(P, G_{\delta})
\end{equation}
Here, $N$ specifies the maximum allowable shift, controlling tolerance for structural variations.

The \textit{CellAlignment} function measures similarity after applying the shift $\delta$, ensuring that legitimate variations (e.g., merged vs.\ split headers) are not penalized.  
To enhance robustness, we flatten content at the index level: all cells in the same row (or column) are concatenated, enabling evaluation to focus on whether semantically related content aligns across structural positions. This index-level evaluation correctly recognizes equivalence in tables with different granularities while still flagging genuine layout errors.

\paragraph{Interpretable Diagnostics.}
Our metrics produce actionable diagnostics \\
(e.g., 89\% content accuracy, 72\% index accuracy with 2-cell shift), offering transparent insight into system strengths and weaknesses. These are complemented by TEDS for hierarchical evaluation.

\paragraph{TEDS Integration.}
We integrate the TEDS metric—correcting bugs in prior implementations—to evaluate hierarchical table structures, capturing relationships that extend beyond cell-level accuracy.

\subsection{Structural Hierarchy Understanding}

Beyond recognizing tables and text, document parsing systems must also capture the hierarchical organization of content. Evaluating this dimension requires metrics that look past surface labels and instead test whether systems maintain consistent, semantically coherent hierarchies across a document.

We map heterogeneous element labels into functional categories, which provides a better evaluation. For example, system labels such as ``title,'' ``sub-title,'' and ``sub-heading'' all map to the functional category ``TITLE,'' enabling semantic-level comparison across different labeling schemes.

VLMs enable semantic-level classification, but reliable \textbf{element consistency} evaluation requires testing stability across similar contexts. We measure whether models assign consistent labels to functionally similar elements using a \textbf{confusion matrix} framework.

\paragraph{Confusion Matrix.}
Let $G$ and $S$ denote the sets of ground-truth and predicted elements, labeled by functions $g$ and $s$ over types $\mathcal{Y}=\{1,\dots,K\}$. A partial matching $M \subseteq G \times S$ pairs aligned elements (e.g., by order or text similarity). To capture both missed detections and spurious predictions, we extend the label set with a \textit{NOMATCH} token, giving $\mathcal{Y}^{+}=\mathcal{Y} \cup \{\text{NOMATCH}\}$. The completed matching $\overline{M}$ pairs all elements, matched or unmatched, and defines the confusion matrix

\[
C_{ij}=\sum_{(x,y)\in \overline{M}}\mathbf{1}[g(x)=i] \cdot \mathbf{1}[s(y)=j], \quad i,j \in \mathcal{Y}^{+}.
\]

Diagonal entries count correct classifications; rows ending in NOMATCH indicate missed elements; columns from NOMATCH indicate false positives.  

\paragraph{Example.}
If the ground truth is $\{\text{title}, \text{text}, \text{text}, \text{figure}\}$ and predictions are $\{\text{title}, \text{table}, \text{text}\}$, the resulting confusion matrix reveals one correct title, one correct text, one missed text, one missed figure, and one spurious table:

\[
\begin{array}{c|ccccc}
\textbf{True $\downarrow$ / Pred $\rightarrow$} 
 & \text{title} & \text{table} & \text{text} & \text{figure} & \text{NOMATCH} \\
\hline
\text{title}  & 1 & 0 & 0 & 0 & 0 \\
\text{table}  & 0 & 0 & 0 & 0 & 0 \\
\text{text}   & 0 & 0 & 1 & 0 & 1 \\ 
\text{figure} & 0 & 0 & 0 & 0 & 1 \\
\text{NOMATCH}& 0 & 1 & 0 & 0 & 0 \\
\end{array}
\]

This unified structure supports computing an $F_{1}$-based consistency score that reflects semantic stability across elements.

\section{Experimental Design and Implementation}

We implement our framework as a modular and extensible system for comparative benchmarking of document parsing systems across diverse datasets. This section describes the experimental setup, system categories, datasets, and empirical results without interpretation, which is provided in the Discussion section.

\subsection{System Categories and Benchmarking}

We evaluated two categories of document parsing systems provided by Unstructured, reflecting both modern generative approaches and traditional IDP pipelines:

\begin{enumerate}
    \item \textbf{VLM-based Partitioner Systems.} Solutions built on VLMs (e.g., OpenAI GPT-5 Mini, Gemini 2.5 Flash, Claude Sonnet 3.7/4), augmented with prompt-tuning and task-specific optimizations for document structure extraction. These represent production-ready deployments rather than raw VLM outputs.  
    \item \textbf{OD-based Solutions.} Conventional IDP pipelines that integrate object detection, OCR, and table extraction modules to produce structured and deterministic outputs. This category includes Unstructured OSS (Open Source Software) and Unstructured's proprietary "hi\_res" (referred to as the "Hi Res" partition strategy in the product) prioritizing stability and accuracy over generative flexibility.  
\end{enumerate}

\subsection{Dataset Composition}

Our evaluation spans two complementary datasets to ensure robust validation across diverse document types and structural challenges:

\begin{itemize}
    \item \textbf{Mini Holistic.} A 314-page handpicked collection of documents, spanning varying complexity levels and layouts, designed to reflect real-world document variability and serve as the primary evaluation corpus.  
    \item \textbf{Industry Documents} A 800-page customer-provided dataset drawn from production use cases, characterized by challenging layouts and domain-specific structures. This serves as an extended validation set to test system robustness and consistency across diverse document types.  
\end{itemize}

\textit{Note: Both datasets are proprietary and not publicly available at this time.}

\subsection{Experimental Results}

We present results across three evaluation dimensions: content fidelity, table structure analysis, and element alignment. For content fidelity, we report both standard and adjusted NED scores for cases where alternative table interpretations affect approximately 2--5\% of pages.

For brevity in tables and figures, we use shortened names: \textbf{Gemini 2.5} refers to Gemini 2.5 Flash, \textbf{Sonnet 3.7/4} refer to Claude Sonnet 3.7 and Claude Sonnet 4 respectively, \textbf{OSS} refers to Unstructured OSS and \textbf{hi\_res} refers to Unstructured's proprietary Hi Res partition strategy.

\textit{Note: VLM-based system performance is contingent on the specific prompts and configurations used in this evaluation, and results represent a snapshot of these systems under our experimental conditions rather than their absolute capabilities.}
\subsubsection{Content Fidelity Results}

Table~\ref{tab:content-fidelity-cross} presents content extraction performance across both datasets. The \textit{Diff} column reports the number of document pages with differing table interpretations, while the \textit{Avg.} column indicates the average score difference observed in those cases. See section 7.2 for a \textit{Diff} example.

\begin{table}[htbp]
\centering
\small
\begin{tabular}{|l|c|c|c|c|c|c|}
\hline
\textbf{System} & \textbf{Adj. NED} & \textbf{NED} & \textbf{Diff} & \textbf{Avg.} & \textbf{T. Added} & \textbf{T. Found} \\
\hline
\multicolumn{7}{|c|}{\textit{Mini Holistic}} \\
\hline
Gemini 2.5 & 0.895 & \textbf{0.891} & 8/314 & 0.130 & 0.063 & \textbf{0.955} \\
GPT-5 Mini & \textbf{0.896} & 0.888 & 13/314 & 0.192 & \textbf{0.043} & 0.929 \\
Sonnet 4 & 0.865 & 0.857 & \textbf{14/314} & 0.188 & 0.068 & 0.936 \\
Sonnet 3.7 & 0.870 & 0.863 & 12/314 & \textbf{0.211} & 0.073 & 0.935 \\
hi\_res & 0.850 & 0.846 & 7/314 & 0.188 & 0.062 & 0.930 \\
OSS & 0.788 & 0.779 & 13/314 & 0.203 & 0.079 & 0.928 \\
\hline
\multicolumn{7}{|c|}{\textit{Industry Documents}} \\
\hline
Gemini 2.5 & 0.883 & 0.879 & 22/800 & 0.157 & 0.072 & 0.956 \\
GPT-5 Mini & 0.873 & 0.868 & \textbf{30/800} & 0.147 & \textbf{0.039} & 0.928 \\
Sonnet 4 & \textbf{0.885} & \textbf{0.882} & 20/800 & 0.139 & 0.069 & \textbf{0.959}  \\
Sonnet 3.7 & 0.859 & 0.855 & 24/800 & 0.126 & 0.085 & 0.949 \\
hi\_res & 0.845 & 0.839 & \textbf{30/800} & 0.170 & 0.048 & 0.944 \\
OSS & 0.806 & 0.798 & 28/800 & \textbf{0.244} & 0.065 & 0.941 \\
\hline
\end{tabular}
\caption{Content fidelity results across both datasets. TokensFound/TokensAdded metrics distinguish system behaviors in content extraction and hallucination control.}
\label{tab:content-fidelity-cross}
\end{table}

\subsubsection{Table Structure Analysis Results}

Table~\ref{tab:table-comprehensive-cross} presents comprehensive table evaluation results. The separation of content and index accuracy enables identification of distinct system capabilities in recognition versus spatial reasoning.

\begin{table}[htbp]
\centering
\footnotesize
\begin{tabular}{|l|c|c|c|c|}
\hline
\textbf{System} & \textbf{Content Acc.} & \textbf{Index Acc.} & \textbf{Detection F1} & \textbf{TEDS}\\
\hline
\multicolumn{5}{|c|}{\textit{Mini Holistic}} \\
\hline
Gemini 2.5 & 0.772 & 0.721 & \textbf{0.930} & 0.809 \\
GPT-5 Mini & 0.775 & 0.706 & 0.870 & 0.788 \\
Sonnet 4 & \textbf{0.833} & 0.748 & 0.903 & \textbf{0.835} \\
Sonnet 3.7 & 0.807 & \textbf{0.774} & 0.920 & 0.814 \\
hi\_res & 0.803 & 0.746 & 0.879 & 0.821 \\
OSS & 0.570 & 0.499 & 0.856 & 0.565 \\
\hline
\multicolumn{5}{|c|}{\textit{Industry Documents}} \\
\hline
Gemini 2.5 & 0.631 & 0.626 & 0.817 & 0.692 \\
GPT-5 Mini & 0.668 & 0.712 & 0.840 & 0.712 \\
Sonnet 4 & \textbf{0.669} & 0.605 & 0.856 & 0.698 \\
Sonnet 3.7 & 0.668 & \textbf{0.680} & \textbf{0.891} & \textbf{0.723} \\
hi\_res & 0.634 & 0.646 & 0.766 & 0.693 \\
OSS & 0.561 & 0.529 & 0.814 & 0.566 \\
\hline
\end{tabular}
\caption{Table structure and content evaluation results across both datasets.}
\label{tab:table-comprehensive-cross}
\end{table}

\subsubsection{Element Alignment Results}

Table~\ref{tab:element-alignment} presents spatial-semantic relationship assessment results, measuring how systems construct and maintain document hierarchies.

\begin{table}[htbp]
\centering
\small
\begin{tabular}{|l|c|}
\hline
\textbf{System} & \textbf{Consistency Level} \\
\hline
\multicolumn{2}{|c|}{\textit{Mini Holistic}} \\
\hline
Gemini 2.5 & 0.352 \\
GPT-5 Mini & 0.377 \\
Sonnet 4 & 0.403 \\
Sonnet 3.7 & \textbf{0.407} \\
hi\_res & 0.305 \\
OSS & 0.195 \\
\hline
\multicolumn{2}{|c|}{\textit{Industry Documents}} \\
\hline
Gemini 2.5 & 0.413  \\
GPT-5 Mini & 0.410 \\
Sonnet 4 & \textbf{0.474} \\
Sonnet 3.7 & 0.471 \\
hi\_res & 0.313 \\
OSS & 0.229 \\
\hline
\end{tabular}
\caption{Element consistency results demonstrating spatial-semantic relationship assessment across document types.}
\label{tab:element-alignment}
\end{table}

\subsection{Summary of Experimental Findings}

The experimental results demonstrate measurable differences across systems and evaluation dimensions:

\begin{itemize}
    \item VLM-based systems (Gemini 2.5, GPT-5 Mini, Sonnet 3.7/4) generally achieve higher content fidelity scores than OD-based approaches (hi\_res, OSS)
    \item Performance rankings vary depending on the specific metric used (content vs. index accuracy, NED vs. adjusted NED)
    \item Cross-dataset consistency varies by system, with some maintaining stable relative performance while others show dataset-dependent variations
    \item The difference between standard and adjusted NED scores affects 2--5\% of pages, with variations ranging from 0.126 to 0.244 average score differences
\end{itemize}

These results provide the empirical foundation for the analysis and interpretation presented in the Discussion section.
\section{Discussion}

This work introduces an interpretation-agnostic evaluation framework for multi-modal generative document parsing systems, motivated by the inadequacy of traditional, deterministic metrics when applied to models capable of producing semantically valid but structurally diverse outputs. Our findings underscore the importance of shifting from purely task-specific measures toward holistic, semantically grounded evaluation. By accounting for representational diversity, spatial flexibility, and hierarchical consistency, the proposed framework offers a principled way to compare systems that may differ in output format or structural interpretation but preserve equivalent meaning.

\subsection{Key Advantages of the Proposed Framework}

Our evaluation across multiple datasets and parsing strategies highlights five major advantages of the proposed framework over conventional approaches:

\begin{enumerate}
\item \textbf{Multi-Dimensional Performance Characterization:} The framework captures system performance across complementary dimensions---content fidelity, spatial reasoning, and structural hierarchy---revealing distinct system strengths. 
In content fidelity, Gemini~2.5 excels with the highest adjusted NED (0.895) and superior token coverage ($\mathrm{TokensFound} = 0.955$), while differences of 0.1\%--0.5\% among top-performing systems are negligible in practical applications. 
GPT~5 Mini consistently demonstrates the lowest hallucination rate ($\mathrm{TokensAdded} = 0.043$ on Mini Holistic, $0.039$ on Industry Documents), though content fidelity remains unstable. 
Sonnet~4 demonstrates superior overall structural understanding compared with other systems, considering both table extraction and element consistency. These distinctions highlight that system choice depends on task priorities rather than a single aggregate score.

\item \textbf{Ranking Corrections:} The framework reveals ranking differences that would be obscured by single-metric evaluation. For example, in the Mini Holistic dataset (Table~\ref{tab:content-fidelity-cross}), Sonnet~4 and Sonnet~3.7 have close unadjusted NED scores (0.857 vs. 0.863). After applying adjusted NED, the gap narrows further (0.865 vs. 0.870), indicating that part of Sonnet~4's apparent deficit was due to benign interpretation differences. More notably, while Gemini~2.5 leads in unadjusted NED (0.891), GPT~5 Mini achieves the highest adjusted NED (0.896), demonstrating how the framework's interpretation tolerance can reveal the true ranking when accounting for valid structural variations. This correction highlights that GPT~5 Mini's superior semantic accuracy was partially masked by traditional metrics that penalized its interpretive diversity. The framework therefore produces rankings that reflect semantic accuracy rather than penalizing models for structurally valid variations.

\item \textbf{Interpretation Tolerance Validation:} The difference between standard and adjusted NED scores (ranging from 0.126 to 0.244 average differences) demonstrates the framework's ability to distinguish genuine errors from benign interpretation variations. Systems showing larger adjustment benefits, such as Sonnet~3.7 (0.211 average difference on Mini Holistic), indicate greater interpretive diversity in their outputs, which traditional metrics would incorrectly penalize.

\item \textbf{Element Alignment as a Reality-Grounded Metric:} Interestingly, OD-based models (hi\_res and OSS) perform closer to VLM systems on alignment than on consistency. This is largely because deterministic pipe-lines handle large, well-defined elements such as tables and figures reliably, narrowing the performance gap. However, their lower consistency levels highlight limited adaptability when faced with ambiguous or nuanced structures. Together, these findings reinforce the importance of element alignment: it captures how systems map heterogeneous labels into coherent hierarchies, even when absolute consistency is elusive.

\item \textbf{Consistency Challenges in Complex Layouts:} Consistency levels are generally low across all systems due to the inherent ambiguity in reading orders and the non-rigid nature of segmentation. Since multiple plausible interpretations often exist for the same document, achieving high consistency scores is inherently difficult. This observation underscores the necessity of evaluation frameworks that tolerate interpretive diversity rather than penalize systems for producing semantically valid but structurally different outputs.

\end{enumerate}

\subsection{Embracing Interpretive Diversity}

A central contribution of this work is its recognition that divergent outputs do not necessarily imply model failure. The framework explicitly treats alternative but semantically equivalent interpretations as valid, aligning evaluation more closely with human judgment. This is especially critical for complex documents containing nested tables, multi-column layouts, or ambiguous reading orders, where multiple hierarchies may faithfully represent the same content.

\subsection{Illustrative Examples}

Appendix~A Section~7.1 demonstrates the value of interpretation-agnostic evaluation through a concrete example. A ground-truth table is encoded as coordinate-based cells, whereas the model outputs a semantically rich HTML table with hierarchical structure and additional descriptive context. Traditional metrics such as TEDS would severely penalize the model (score $0.34$), falsely suggesting a significant error. In contrast, cell content accuracy correctly recognize that the semantic content is preserved (score $0.68$), validating the model's output. This case study highlights how the proposed framework distinguishes genuine errors from acceptable structural variance, leading to fairer and more informative assessments. Section~7.2 further illustrates how human and model outputs can follow very different reading paths while remaining semantically equivalent. Traditional NED penalizes the model heavily (score $0.36$), while adjusted NED correctly recognizes semantic fidelity (score $0.91$). 
This example reinforces that rigid, format-specific metrics risk misclassifying valid outputs as errors.

\section{Conclusions and Future Work}

By providing a unified, format-agnostic approach that balances rigor with interpretive tolerance, our framework enables more meaningful benchmarking of generative document systems. These insights can guide system selection, hyperparameter tuning, and architecture design for real-world deployments. Future work will explore incorporating human preference models to weight interpretive diversity, extending evaluation to multi-modal fusion tasks, and integrating our metrics into public leaderboards to encourage adoption of semantically aware evaluation practices.

\bibliographystyle{plain}
\bibliography{bibliography}

\section{Appendix A - Examples}
\label{sec:appendix-examples}

\subsection{Multi-dimensional insights: example of cell content accuracy vs TEDS}
\label{sec:appendix-cell-content-accuracy}

The example below shows the ground truth in figure~\ref{fig:table-gt} and the prediction in figure~\ref{fig:table-html}. We can see that the prediction kept all the content into a single cell that was arranged into different subsections and paragraphs. This is still a valid interpretation of the table. 

Using a single metric that combines structure and content without enough flexibility, we see that it would look as if the prediction missed quite a significant amount of information from the ground truth (TEDS of 0.34), while the cell accuracy metric was able to adapt the interpretation of the table (score of 0.68).

\begin{figure}[htbp]
\begin{verbatim}
[
  {
    "type": "Table",
    "text": [
      {
        "x": 0, "y": 0, "w": 1, "h": 1,
        "content": "Dabblers"
      },
      {
        "x": 0, "y": 1, "w": 1, "h": 1,
        "content": "15% of 50-plus Gamers"
      },
      {
        "x": 0, "y": 2, "w": 1, "h": 1,
        "content": "Gaming is not an integral aspect of the Dabbler’s
        life; they play infrequently to pass the time and relieve
        boredom. Gaming is not perceived as hugely beneficial to them
        as they age, and they have no desire to game more than they
        already do. When playing, they’re doing it mostly alone,
        playing card, tile, and puzzle games on their PCs or phones."
      },
      ...
]
\end{verbatim}
\caption{Example of table ground truth. Each cell has a set of coordinates withing the table (x, y) and an annotation of units of length each cell takes (w, h), and its text content.}
\label{fig:table-gt}
\end{figure}

\begin{figure}[htbp]
\begin{verbatim}
<table>
  <tbody>
    <tr>
      <td>
        <h2>Dabblers</h2>
        <p><span>15%</span><br/><p>of 50-plus Gamers</p></p>
        <img alt="A middle-aged Black man looking at his mobile
        device" class="Image"/>
        <p>
          <p>Gaming is not an integral aspect of the Dabbler's life;
          they play infrequently to pass the time and</p>
          <span>relieve boredom</span>
          <p>. Gaming is not perceived as hugely beneficial to them
          as they age, and they have</p>
          <span>no desire to game more than they already do</span>
          <p>. When playing, they're doing it mostly alone, playing</p>
          <span>card, tile, and puzzle games</span>
          <p>on their PCs or phones.</p>
        </p>
      </td>    
      ...</tr>
  </tbody>
</table>
\end{verbatim}
\caption{Example table extracted by Anthropic Claude Sonnet 3.7 in HTML format.}
\label{fig:table-html}
\end{figure}

\subsection{Adj. NED Example: Semantic Equivalence in Different Reading Paths}
\label{sec:reading_paths_example}

This section demonstrates how our model can generate different reading paths than human interpretations while maintaining semantic equivalence. The example shows two Wikimedia navigation tables where the model's HTML output differs structurally from the human-labeled JSON format, yet both representations convey the same semantic information.

\begin{figure}[htbp]
    \centering
    \includegraphics[width=0.8\textwidth]{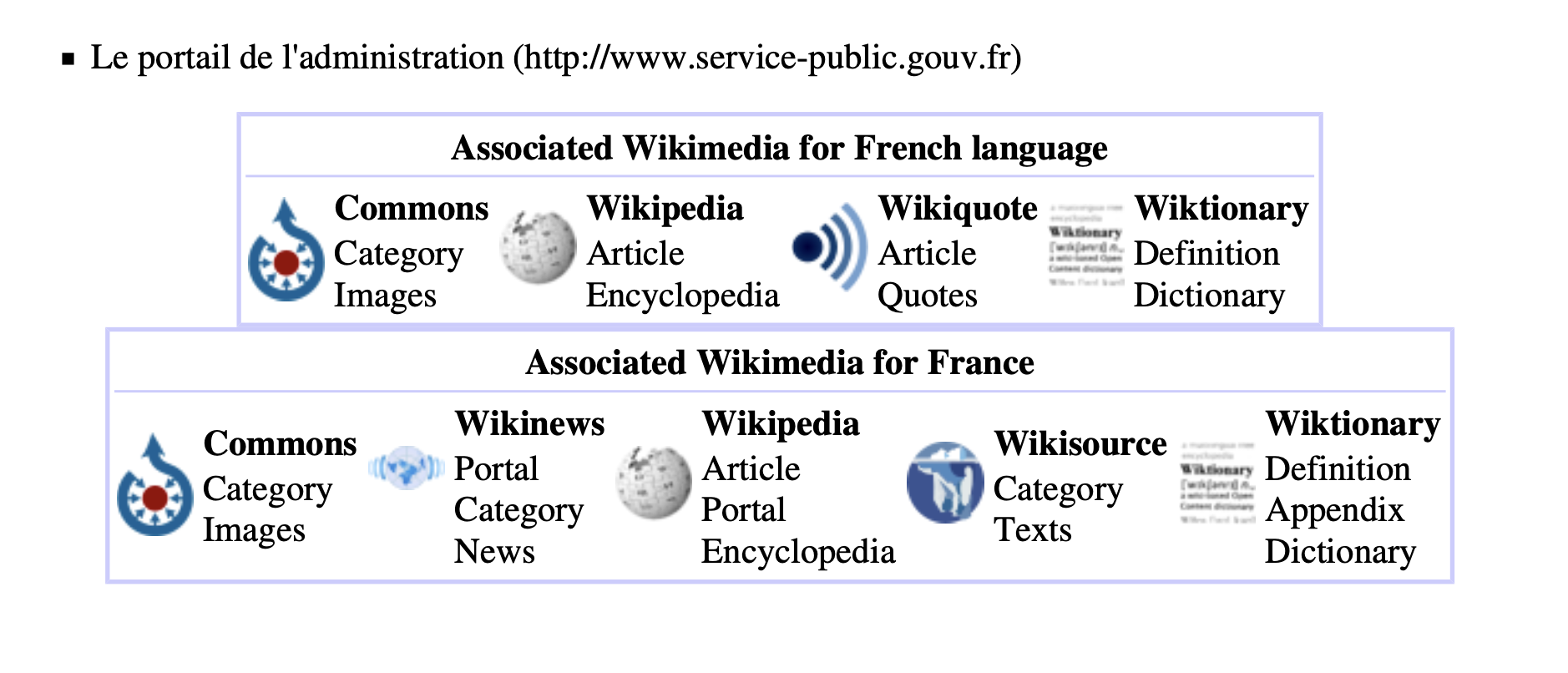}
    \caption{Different reading paths}
    \label{fig:your_label}
\end{figure}

\subsubsection{Human Annotation}
The human annotators structured the content as coordinate-based table cells with explicit positioning:

\begin{figure}[htbp]
\begin{verbatim}
{
"type": "Table",
"text": [
  {
    "x": 0, "y": 0, "w": 4, "h": 1,
    "content": "Associated Wikimedia for French language"
  },
  {
    "x": 0, "y": 1, "w": 1, "h": 1,
    "content": "Commons Category Images"
  },
  {
    "x": 1, "y": 1, "w": 1, "h": 1,
    "content": "Wikipedia Article Encyclopedia"
  },
  {
    "x": 2, "y": 1, "w": 1, "h": 1,
    "content": "Wikiquote Article Quotes"
  },
  {
    "x": 3, "y": 1, "w": 1, "h": 1,
    "content": "Wiktionary Definition Dictionary"
  }
]
}
\end{verbatim}
\caption{Ground truth in cells format. Each cell has a set of coordinates withing the table (x, y) and an annotation of units of length each cell takes (w, h), and its text content.}
\end{figure}

\subsubsection{Model Prediction}
The model generated the following HTML representation:

\begin{figure}[htbp]
\begin{verbatim}
<table class="Table">
  <thead>
    <tr>
      <th colspan="4">Associated Wikimedia for French language</th>
    </tr>
  </thead>
  <tbody>
    <tr>
      <td class="Project">
        <img class="Logo"/>
        <div class="Info">
          <div class="Name">Commons</div>
          <div class="Type">Category</div>
          <div class="Desc">Images</div>
        </div>
      </td>
      <td class="Project">
        <img class="Logo"/>
        <div class="Info">
          <div class="Name">Wikipedia</div>
          <div class="Type">Article</div>
          <div class="Desc">Encyclopedia</div>
        </div>
      </td>
      <td class="Project">
        <img class="Logo"/>
        <div class="Info">
          <div class="Name">Wikiquote</div>
          <div class="Type">Article</div>
          <div class="Desc">Quotes</div>
        </div>
      </td>
      <td class="Project">
        <img class="Logo"/>
        <div class="Info">
          <div class="Name">Wiktionary</div>
          <div class="Type">Definition</div>
          <div class="Desc">Dictionary</div>
        </div>
      </td>
    </tr>
  </tbody>
</table>
\end{verbatim}
\caption{Table prediction in HTML format}
\end{figure}

This example illustrates a key finding in our evaluation: while the model's reading path differs significantly from human interpretation, both representations are semantically equivalent. The human annotation uses a coordinate-based approach with explicit x, y positioning and cell dimensions, treating each piece of content as discrete positioned elements. In contrast, the model generates standard HTML table markup with hierarchical structure, incorporating visual elements (logos) and using semantic HTML tags like \texttt{<p>} for text organization.

\subsubsection{Analysis}

Despite these structural differences, both representations preserve the same essential information: the table headers, the organization of Wikimedia projects, and the descriptive text for each service. The model's approach actually provides additional semantic richness by including visual context (logo descriptions) and proper HTML semantics, demonstrating that alternative reading paths can enhance rather than diminish content representation while maintaining semantic fidelity. In this case, standard NED assigns a low score ($0.36$), harshly penalizing the model’s structurally different output, whereas the adjusted NED yields a much higher score ($0.91$), correctly capturing the preserved semantic fidelity.

\end{document}